\documentclass{article} 
\usepackage{iclr2027_conference,times}

\usepackage{amsmath,amsfonts,bm}

\def\eqref#1{equation~\ref{#1}}

\def\1{\bm{1}}

\DeclareMathAlphabet{\mathsfit}{\encodingdefault}{\sfdefault}{m}{sl}
\SetMathAlphabet{\mathsfit}{bold}{\encodingdefault}{\sfdefault}{bx}{n}

\usepackage{hyperref}
\usepackage{url}
\usepackage{graphicx}
\usepackage{booktabs}
\usepackage{tabularx,multirow}
\usepackage{xcolor,colortbl}

\title{WEAVE: Compressing State Changes Along Native Evolution in Linear Attention LLMs}

\author{
\parbox[t]{0.95\textwidth}{
\raggedright
Zedong~Liu$^{1,2}$ \quad
Xinyang~Ma$^{1,2}$ \quad
Man~Liu$^{1,2}$ \quad
Bing~Lu$^{1,2}$ \quad
Chen~Yuhan$^{1,2}$ \quad
Dingwen~Tao$^{2}$ \quad
Guangming~Tan$^{2}$ \\[6pt]
\normalfont
$^{1}$University of Chinese Academy of Sciences \\
$^{2}$Institute of Computing Technology, Chinese Academy of Sciences
}
}

\iclrfinalcopy
\begin{document}
\pagestyle{plain}
\thispagestyle{plain}

\maketitle

\begin{abstract}
A new generation of hybrid linear-attention models replaces much of the
growing KV cache with fixed-size recurrent states. Yet agent serving must retain many of these
states as checkpoints for prefix reuse across turns and branches, creating
substantial GPU-memory pressure and costly prefix recomputation under load. 
We present WEAVE, a training-free approach that encodes state changes from a shared anchor and efficiently constructs these compact representations from the model’s native updates. WEAVE combines shared-anchor residuals, recurrence-driven encoding from native model updates, and periodic compaction with fused GPU kernels for efficient online serving.
We evaluate WEAVE on Qwen3.5-35B-A3B and Kimi-Linear-48B across reasoning, long-context, and agentic workloads. WEAVE compresses recurrent checkpoints by up to $4.26\times$ while maintaining task quality and outperforming INT8 and INT4 quantization on all six benchmarks. Integrated into vLLM's native prefix-cache path, the reduced state footprint lowers prefix eviction and recomputation under memory pressure, yielding up to $6.96\times$ lower TTFT and $3.14\times$ higher token goodput under SLO constraints.
\end{abstract}

\section{Introduction}
\label{sec:intro}

Recent open-source models such as Qwen3.8 and Kimi K3 have demonstrated
strong capabilities on long-context and agentic workloads, while adopting
a new class of \emph{hybrid linear-attention architectures}
\citep{yang2025gdn,kimiteam2025kimi}.
These models replace roughly three quarters of their full-attention layers
with linear-attention mechanisms such as Gated DeltaNet (GDN) and
Kimi Delta Attention (KDA).
Unlike conventional KV caches that grow with sequence length, these
linear-attention layers summarize historical context into fixed-size
\emph{recurrent states}, substantially reducing the cache growth and
memory-bandwidth pressure exposed during long-context decoding.

However, a fixed-size current state does not eliminate the storage problem.
Agent workloads repeatedly reuse prefixes across turns and branches: 
production agent traces report a median input length of
142K tokens and a prefix-cache hit rate above 96\%
\citep{agentx_trace}.
Serving systems therefore retain recoverable
\emph{recurrent-state checkpoints} to reuse as much of each prefix as possible.
These checkpoints are themselves expensive.
For Qwen3.5-35B, one state checkpoint occupies roughly the same
storage as 3.2K tokens of its full-attention KV cache.
Consequently, denser checkpointing improves prefix-reuse granularity but
consumes more HBM capacity.
Under high load, this pressure forces recurrent checkpoints and KV blocks
to be evicted. Long prefixes must then be recomputed, increasing time-to-first-token (TTFT) and reducing the concurrency the system can sustain.

Compressing recurrent-state checkpoints is therefore a natural solution.
Prior work such as KIVI and KVQuant
\citep{liu2024kivi,hooper2024kvquant}, has shown that quantization can substantially reduce
the storage cost of full-attention KV caches while preserving model quality.
Recurrent states, however, are not collections of independent token-indexed K/V vectors. They are matrix-valued memories that
are continuously transformed and overwritten during forward execution.
In our experiments, INT8 quantization can still preserve model behavior
well, whereas further reducing states to INT4 causes
reconstruction error and downstream quality degradation rapidly.
Treating recurrent state as an ordinary dense tensor therefore
encounters a sharp rate--distortion bottleneck.

The key opportunity comes from how linear-attention states are generated.
GDN and KDA do not construct each new state from scratch.
Instead, they update an existing state by transforming \textit{prior memory} and injecting a sequence of \textit{structured writes}~\citep{nazari2026statereduction}.
We find that after removing this predictable evolution, state changes
across checkpoints exhibit pronounced low-dimensional structure:
write directions are repeatedly reused over long spans, while the effective
rank of the residual remains low and grows slowly with checkpoint span.
This suggests encoding \emph{state changes} along the trajectory rather than compressing each full state independently, creating opportunities for lower storage cost than conventional low-bit quantization.


\begin{figure}[t!]
  \centering
  \includegraphics[width=1.0\linewidth]{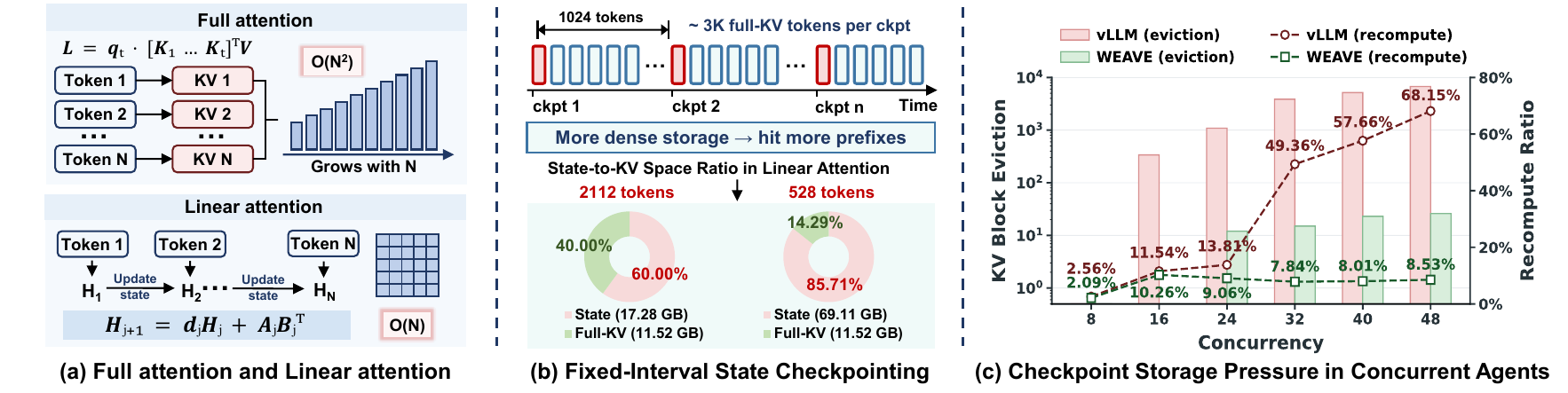}
\caption{\textbf{Why recurrent-state checkpoints become a memory bottleneck.}
(a) Linear attention replaces growing KV caches with fixed-size states.
(b) Finer checkpoints improve prefix reuse but increase memory pressure.
(c) Under concurrency, this causes KV eviction and recomputation.}
  \label{fig:motivation}
  \end{figure}

\paragraph{Challenges.}Turning this observation into an online codec is nevertheless non-trivial.
A straightforward chained delta representation makes each checkpoint
recursively dependent on its predecessor: restoring a prefix may require replaying a long chain, while losing an intermediate checkpoint can break recovery.
Materializing and factorizing each dense residual with SVD is also too costly for online serving.
The challenge is: whether we can achieve \emph{high compression} while preserving \emph{model quality} and keeping checkpoint construction and recovery \emph{efficient}.

We present \textbf{WEAVE}, a recurrent-state compression method that
compresses state changes along the model's native evolution rather than
treating every checkpoint as an independently generated dense tensor.
Our contributions are as follows:
\begin{itemize}
    \item We systematically characterize the temporal evolution of recurrent
    states in hybrid linear-attention LLMs and show that state changes exhibit
    substantially stronger low-rank structure than full states.
    Based on this observation, WEAVE uses a
    shared anchor and stores only compact residual factors for subsequent checkpoints.

    \item We design a recurrence-driven encoder that directly consumes structured updates already produced by linear attention. It maintains the compressed representation during inference without materializing dense deltas or performing expensive post-hoc factorization, achieving up to $4.26\times$ compression with substantially higher reconstruction quality than aggressive low-bit quantization.

    \item We co-design WEAVE with the serving system through
    GPU-efficient checkpoint construction, and integration into vLLM's native
    prefix-cache path. Fused Triton kernels keep online encoding
    efficient, achieving low cost and at most 3.51\% latency interference.

    \item We evaluate WEAVE on two representative
    models and six long-context, reasoning, and agentic workloads.
    WEAVE achieves a stronger compression--quality tradeoff than all evaluated
    baselines and translates the saved state capacity into end-to-end
    serving benefits, reducing TTFT by $2.84\times$ on average and up to $6.96\times$,
    while improving token goodput by $1.76\times$ on average and up to
    $3.14\times$.

\end{itemize}


\begin{figure}[t!]
  \centering
  \includegraphics[width=1.0\linewidth]{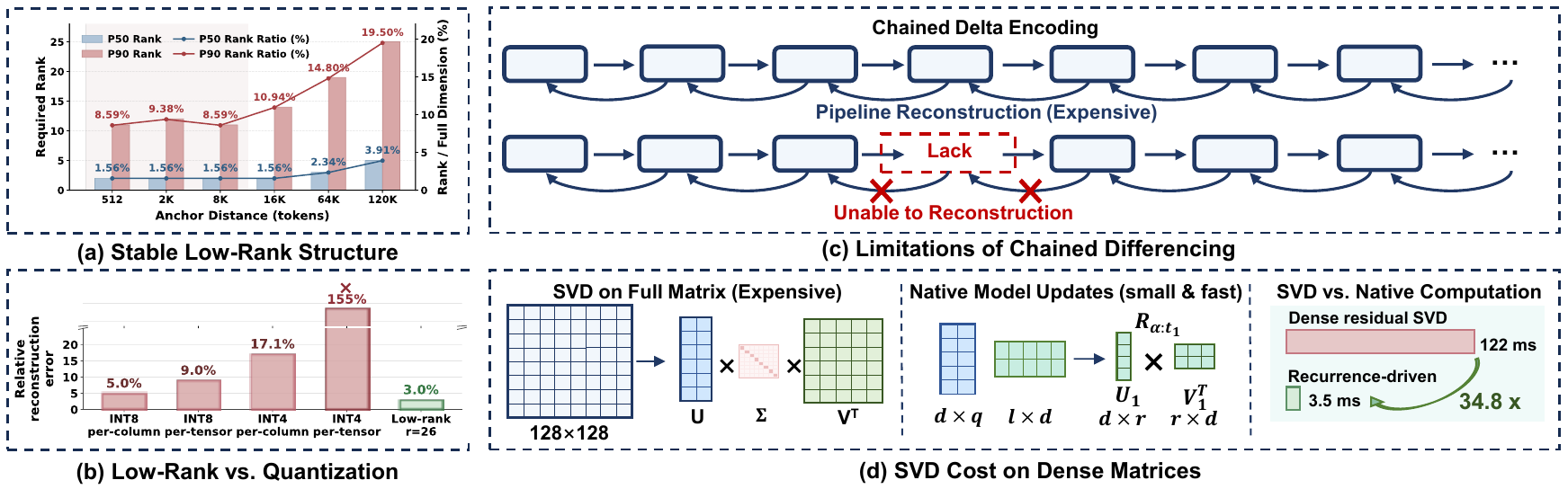}
\caption{\textbf{Motivation for recurrent-state compression.}
(a) Shared-anchor residuals remain low rank over long distances.
(b) Residual encoding incurs less reconstruction error than direct state quantization.
(c) Chained deltas create fragile, serial recovery dependencies.
(d) Dense residual factorization is too costly for online checkpointing.}
  \label{fig:observation}
  \end{figure}
  

\section{Recurrent States and Persistent Checkpointing}
\label{sec:background}

\subsection{Recurrent States in Hybrid Linear Attention}
\label{sec:recurrent_state}

Conventional full attention explicitly preserves past tokens in a KV cache.
As the context grows, its storage increases linearly with sequence length and
the attention computation repeatedly accesses these historical K/V vectors.
Linear attention instead summarizes the history into a fixed-size recurrent
state~\citep{katharopoulos2020transformers}.
As illustrated in Figure~\ref{fig:motivation}(a), only the current state is
needed during sequential execution, largely removing the context-dependent
cache growth of linear-attention layers.
Modern hybrid architectures combine both mechanisms: a small fraction of
layers retain token-indexed KV caches, while the remaining layers maintain
fixed-size recurrent states.

Different linear-attention mechanisms evolve these states differently, but
they share the same basic structure: existing memory is first transformed and
then updated with newly computed information.
For example, the chunk-level transitions of GDN and KDA can be written as
\begin{equation}
    H_{j+1}=d_jH_j+A_jB_j^\top
    \qquad\qquad
    H_{t+1}=H_tD_t+W_t,
    \label{eq:gdn_kda_transition}
\end{equation}
respectively.
In GDN, $d_j$ controls how the previous state is retained and
$A_jB_j^\top$ is a structured write from the current chunk.
KDA applies channel-wise transport on the right through the diagonal matrix $D_t$,
followed by the delta-rule innovation $W_t$.
Despite their different update rules, both construct the next state by
\emph{evolving} an existing one rather than generating an independent dense
matrix.
This temporal structure is central to the compression opportunities explored
by WEAVE.

\subsection{Checkpoint Compression: Opportunity and Challenges}
\label{sec:checkpoint_motivation}

To reuse shared prefixes, serving systems retain recurrent-state checkpoints at regular token intervals, as shown in Figure~\ref{fig:motivation}(b).
Shorter intervals reduce recomputation after a prefix hit but store more states.
This tradeoff is expensive because each checkpoint is large.
In our Qwen3.5-35B serving setup, reducing the checkpoint interval from
2,112 to 528 tokens increases recurrent-state storage from 17~GB to 69~GB,
consuming about 85\% of the available cache memory
(Figure~\ref{fig:motivation}(b)).
Under concurrent load, this pressure increases cache evictions and recomputation (Figure~\ref{fig:motivation}(c)).

Quantization helps only to a point. Unlike token-indexed K/V vectors, a recurrent state is a coupled matrix that
is repeatedly consumed and modified by subsequent transitions.
INT8 introduces modest reconstruction error, whereas INT4 causes the error to rise sharply (Figure~\ref{fig:observation}(b)).
Simply applying lower bit-widths therefore reaches a poor
rate--distortion regime before achieving the desired memory reduction ~\citep{hatamizadeh2026gdnet2}.

Checkpoint residuals offer a different compression opportunity.
Across anchor distances up to 8K tokens, the median rank required to capture
95\% of residual energy is only 2 out of 128 dimensions, and even at a
122K-token distance, the median rank remains 5 (as shown in Figure~\ref{fig:observation}(a)).
This low-rank structure persists over long spans, but exploiting it online raises two challenges.

\textbf{Challenge 1: Recovery dependencies.} Chained delta encoding makes each checkpoint depend on earlier ones. Recovery may require replaying an inefficient long chain, and losing an intermediate delta can invalidate later checkpoints (Figure~\ref{fig:observation}(c)).

\textbf{Challenge 2: Encoding cost.} Directly forming and factorizing each dense residual avoids such chains, but dense SVD takes 122~ms for 32 recurrent heads in our online test (Figure~\ref{fig:observation}(d)). The codec must capture the residual's low-rank structure without adding this cost to the serving path.

\begin{figure}[t!]
  \centering
  \includegraphics[width=1.0\linewidth]{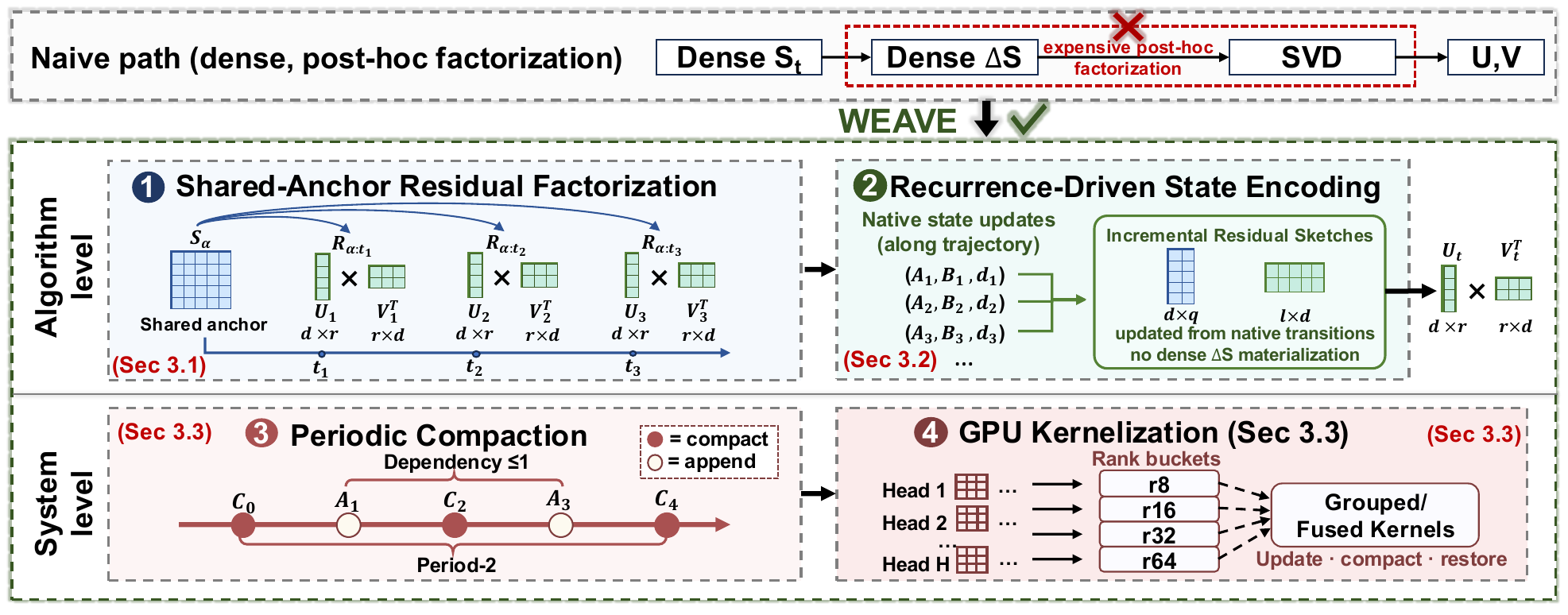}
  \caption{\textbf{Overview of WEAVE.}
    WEAVE first represents recoverable states using a shared anchor and
    compact residual factors.
    It then constructs those factors directly from recurrence-generated state
    updates, avoiding dense post-hoc factorization.
    Periodic append/compact checkpoints bound reconstruction dependency, while
    grouped and fused GPU kernels execute update, compaction, and restoration
    efficiently across recurrent heads.}
  \label{fig:weave}
  \end{figure}

\section{WEAVE}
\label{sec:method}

WEAVE does not compress recurrent states as independent dense tensors.
Instead,  It follows their native evolution and encodes only low-rank changes from a shared reference.
The design has three steps.
First, multiple checkpoints share an anchor and store only compact residuals.
Second, WEAVE constructs these residuals directly from structured updates
already produced by linear attention, avoiding post-hoc dense
factorization.
Finally, periodic compaction and GPU kernelization make the codec practical
for online prefix caching.

\subsection{Shared-Anchor Residual Factorization}
\label{sec:anchor}

\paragraph{Shared anchors instead of delta chains.}
The simplest baseline quantizes every recurrent state independently.
It supports random checkpoint recovery but ignores temporal redundancy.
Its rate--distortion also deteriorates quickly at aggressive precision:
in our measurements, INT4 reaches up to 155\% relative reconstruction
error.

Adjacent-state delta coding better exploits temporal locality,
$H_t=H_{t-1}+\Delta_t$, but creates a growing dependency chain.
Recovering a target state requires serially replaying preceding deltas, and
losing one intermediate object invalidates later checkpoints.
WEAVE instead uses a \emph{shared anchor}.
Checkpoints within the same span share an anchor $H_a$, while each child
stores only a compact residual.
This amortizes common state information while supporting direct recovery from
the anchor.

For checkpoints sharing an anchor $H_a$, WEAVE writes
\begin{equation}
H_t=\bar{H}_{a:t}+R_{a:t},
\qquad
R_{a:t}\approx U_tV_t^\top ,
\label{eq:shared_anchor}
\end{equation}
where $\bar{H}_{a:t}$ is the model-specific anchor reference: it remains
fixed for GDN and follows channel-wise transport for KDA.
The recovered checkpoint is therefore
$\hat{H}_t=\bar{H}_{a:t}+U_tV_t^\top$.
The reference captures information shared across checkpoints, while
$U_t,V_t$ encode the change specific to the target prefix.
Its exact form depends on the recurrence structure.

\paragraph{State changes are low rank.}
State changes exhibit strong temporal structure. For GDN, a chunk transition has the form
$H_{j+1}=d_jH_j+A_jB_j^\top$:
the existing state is transported, then modified by a structured write.
Across real workloads, these writes repeatedly reuse a small set of
directions.
Less than 10\% of the state dimension captures roughly 95\% of the residual
energy, and the effective rank grows slowly with checkpoint span.
This makes residual factors more compact than direct state quantization.

Compressibility also varies across recurrent heads.
WEAVE assigns a static stored rank to each $(\text{layer},\text{head})$
offline and maps it to a small set of hardware-friendly buckets, e.g.,
$r\in\{8,16,32,64\}$.
Runtime rank selection only requires a table lookup.
The resulting factors are further row-wise quantized and packed.
For the few sensitive heads that exceed the target error bound, WEAVE uses
bounded repair or exact fallback instead of increasing the rank of every
head.

Shared-anchor factorization defines \emph{what to store}.
It does not yet explain how to build the factors cheaply online.
Materializing $R_{a:t}$ and factorizing it at every checkpoint can still impose substantial overhead on the online path.
We address this next.

\subsection{Recurrence-Driven State Encoding}
\label{sec:encoding}

\paragraph{Encode along the evolution.}
A direct low-rank encoder still follows
$H_t\!\rightarrow\!R_t\!\rightarrow\!\mathrm{Factorize}(R_t)
\!\rightarrow\!(U_t,V_t)$.
For a $128\times128$ residual, dense decomposition costs
$O(128^3)$ operations.
In our online test, this path takes 995\,ms per checkpoint, which would impose substantial overhead on inference.
The representation is useful, but the execution model is not.

Linear attention already computes how the state changes.
For GDN, the native forward pass exposes the transition factors
$A_j,B_j$ and decay $d_j$.
WEAVE consumes these quantities directly rather than inferring the same
structure after the dense state has been produced.

For GDN, the anchor reference remains fixed,
$\bar{H}_{a:j}=H_a$. The residual therefore follows
\begin{equation}
R_{a:j+1}
=
d_jR_{a:j}
+(d_j-1)H_a
+A_jB_j^\top .
\label{eq:gdn_residual}
\end{equation}
Maintaining $R_{a:j}$ still requires a full matrix. Appending every
$A_j,B_j$ instead makes the factor rank grow without bound. WEAVE does not
maintain the final $U,V$ as the state evolves. It keeps only fixed-width
projections, postponing factor construction until a checkpoint is needed.

\paragraph{Exact sketch recurrence.}
For each rank bucket, WEAVE fixes two projection matrices $\Omega$ and $\Psi$
and maintains $Y_j=R_{a:j}\Omega$ and $Z_j=\Psi^\top R_{a:j}$.
Projecting Equation~\ref{eq:gdn_residual} gives
\begin{equation}
\begin{aligned}
Y_{j+1}
&=d_jY_j+(d_j-1)H_a\Omega+A_j(B_j^\top\Omega),\\
Z_{j+1}
&=d_jZ_j+(d_j-1)\Psi^\top H_a+(\Psi^\top A_j)B_j^\top .
\end{aligned}
\label{eq:gdn_sketch}
\end{equation}

Each chunk now updates only low-dimensional sketches and the skinny factors
already produced by the model.
No dense residual is formed.
Importantly, Equation~\ref{eq:gdn_sketch} is an algebraic projection of
the exact residual recurrence.
Approximation arises from finite sketch width and final rank truncation.

WEAVE converts the sketches to final factors only when a checkpoint is
published:
\begin{equation}
Q=\operatorname{orth}(Y),
\qquad
C\approx(\Psi^\top Q)^\dagger Z,
\qquad
R\approx QC\approx U_rV_r^\top .
\label{eq:sketch_to_factor}
\end{equation}
Dense factorization is thus replaced by bounded computation on small
projected matrices.
At checkpoint finalization, WEAVE recovers the factors using shifted
Cholesky-QR, regularized least squares, and a fixed number of covariance
subspace iterations~\citep{tropp2017practical}.
During accumulation, WEAVE keeps a working rank slightly larger than the
stored rank so that new write directions are not discarded too early.
Only the final checkpoint is truncated to the stored rank.
The resulting $U_r$ and $V_r$ factors are stored in INT8 to further reduce the checkpoint footprint.

\paragraph{Transport-aware extension to KDA.}
For KDA, the native chunk update is
$H_{j+1}=H_jD_j+L_jK_j^\top$, where the diagonal $D_j$ applies
channel-wise transport on the right.
Let
$P_{a:t}=\prod_{j=a}^{t-1}D_j$ and
$\bar{H}_{a:t}=H_aP_{a:t}$.
The corresponding residual evolves as
\begin{equation}
R_{a:j+1}=R_{a:j}D_j+L_jK_j^\top .
\label{eq:kda_residual}
\end{equation}
Rather than materializing $R_{a:j}$, WEAVE maintains
$R_{a:j}\approx F_jG_j^\top$ with bounded width. Each native update forms
\begin{equation}
\widetilde F=[F_j,L_j],\qquad
\widetilde G=[D_jG_j,K_j],
\label{eq:kda_factor_update}
\end{equation}
and truncates only the resulting factor-space core back to the working rank.
At a checkpoint boundary, WEAVE truncates to the stored rank and quantizes
the final factors. The diagonal transport is stored as its accumulated
vector. Appendix~\ref{app:kda} gives the full encoding and recovery procedure.

\subsection{Periodic Compaction and System Co-Design}
\label{sec:kernel}

\paragraph{Periodic compaction.}
Recurrence-driven encoding removes dense factorization, but materializing a
fully compact checkpoint at every fine-grained boundary is unnecessary.
WEAVE separates lightweight state accumulation from checkpoint compaction and
periodically folds accumulated changes into self-contained residual factors.
If compaction runs every $p$ child publications, its amortized cost is
$T_{\mathrm{acc}}+T_{\mathrm{compact}}/p$.
Smaller $p$ shortens recovery dependency, while larger $p$ better amortizes
compaction. As illustrated in Figure~\ref{fig:weave}, WEAVE uses $p=2$:
the intermediate boundary keeps only a bounded one-step dependency on the
latest compact checkpoint, balancing recovery depth and online cost.

\paragraph{GPU kernelization.}
Low-dimensional operations can still be dominated by small GEMMs and kernel
launches when executed independently across hundreds of recurrent heads.
WEAVE groups heads by stored rank and processes each bucket in batch.
Head-wise extraction, sketch updates, quantization, packing, and restoration
use grouped or fused Triton kernels, while Cholesky factorization and
triangular solves use batched GPU primitives. This division keeps irregular factor-space operations in fused kernels while mapping numerically sensitive linear algebra to optimized GPU libraries. Checkpoint encoding and periodic compaction are scheduled asynchronously
with model execution, so neither synchronously blocks token generation. We evaluate encoder cost against dense factorization under matched
all-head coverage in Section~\ref{ablation}. WEAVE cuts checkpoint construction time by $94\times$ relative to dense SVD.
Across both models, the full codec adds only 0.7–1.6\% JCT overhead and 2.6–3.4\% TTFT overhead at concurrency 16, including repair and fallback. 

\paragraph{vLLM integration.}
We integrate WEAVE into the native prefix-cache path of vLLM v0.27.1.
Each compressed checkpoint remains bound to vLLM's native prefix hash, exact
token boundary, parent anchor, cache group, and TP shard. It is published only
after its dependencies and restore metadata are ready, and a restored state is
validated before atomically replacing the live recurrent state. On a prefix hit, WEAVE restores the compressed checkpoint into the native
recurrent-state buffer and validates the restored state before atomically
replacing the live state.
Prefix hashing, scheduling, and full-attention KV management remain
unchanged.
WEAVE therefore acts as a drop-in compressed backend, letting the same cache retain more reusable prefixes.


\section{Evaluation}
\label{sec:evaluation}

We evaluate whether WEAVE preserves model behavior under repeated checkpoint
reuse, whether it improves online serving under memory pressure, and which
design account for its quality and cost.

\subsection{Experimental Setup}
\label{sec:eval_setup}

\paragraph{Models and workloads.}
We evaluate two hybrid linear-attention models: Qwen3.5-35B-A3B with Gated
DeltaNet and Kimi-Linear-48B with Kimi Delta Attention. Unless stated
otherwise, experiments use tensor parallelism across two RTX PRO 6000 GPUs
with BF16 model execution. Six benchmarks: MMLU and AIME2026 for
reasoning, RepoQA and QMSum for long-context understanding, and SWE-bench
Verified and AppWorld for agentic execution~\citep{jimenez2024swebench}.

\paragraph{Checkpoint reuse.}
Our quality tests follow the checkpoint lifecycle during inference rather
than compressing a single state offline. We encode checkpoints at multiple
boundaries along each inference trajectory. To simulate a prefix-cache hit,
we restore the recurrent state from a compressed checkpoint and continue
inference from the restored state. The continued trajectory can produce
further checkpoints, so any error introduced at restoration also enters
subsequent state updates. All methods use the same checkpoint boundaries and
continuation inputs.

\paragraph{Baselines and accounting.}
We compare WEAVE with dense recurrent-state checkpoints,
per-column INT8 and INT4 quantization.
We also evaluate direct SVD of dense residuals as an offline reference for
representation quality and encoding cost; it is not used as an online serving
baseline. Compression ratios use the physical size of
recoverable checkpoints, including amortized anchor storage, residual
factors, quantization metadata, convolution states, and repair payloads.
Serving comparisons replay the same request streams with the same scheduler,
prefix-cache policy, and GPU memory budget, changing only the recurrent-state
checkpoint representation.

\subsection{Compression Quality}

\paragraph{Downstream tasks.}
Table~1 shows that full-state quantization trades accuracy for
checkpoint size: INT8 causes modest score reductions on most tasks,
while INT4 degrades much more sharply. The largest relative loss is on
Kimi-Linear-48B's SWE-bench Verified, where INT4 falls from 28.66 to
15.12, a 47.2\% reduction from the dense baseline. WEAVE scores
27.98 on the same task while compressing checkpoints by
$4.12\times$, compared with $3.26\times$ for INT4. On
Qwen3.5-35B-A3B, WEAVE reaches $4.26\times$ compression and
similarly stays close to the dense baseline across the evaluated
tasks. The quality advantage holds when compressed checkpoints
are restored repeatedly during inference, rather than only once
in isolation.


\begin{table}[t]
    \centering
    \caption{\textbf{Downstream task quality and recurrent-state compression.}}
    \label{tab:quality}
    \vspace{4pt}
    \fontsize{8.5}{9.5}\selectfont
    \begingroup
    \setlength{\tabcolsep}{2pt}
    \renewcommand{\arraystretch}{1.08}
    \definecolor{weaverow}{RGB}{236,248,236}
    \begin{tabularx}{\linewidth}{l *{6}{>{\centering\arraybackslash}X} c}
        \toprule
        \multirow{2}{*}[-3.5pt]{\textbf{Method}}
        & \multicolumn{2}{c}{\textbf{Reasoning}}
        & \multicolumn{2}{c}{\textbf{Long-context}}
        & \multicolumn{2}{c}{\textbf{Agentic-bench}}
        & \multirow{2}{*}[-3.5pt]{\shortstack{\textbf{Compression}\\\textbf{Ratio}}} \\
        \cmidrule(lr){2-3}\cmidrule(lr){4-5}\cmidrule(lr){6-7}
        & \textbf{MMLU} & \textbf{AIME2026} & \textbf{RepoQA}
        & \textbf{QMSum} & \textbf{SWE} & \textbf{AppWorld} & \\
        \midrule
        \multicolumn{8}{l}{\textbf{Qwen3.5-35B-A3B}} \\[-2pt]
        Native         & 84.60 & 53.33 & 90.91 & 22.40 & 34.10 & 44.25 & $1.00\times$ \\
        INT8           & 82.32 & 46.67 & 88.89 & 22.26 & 32.44 & 42.69 & $1.89\times$ \\
        INT4           & 80.16 & 46.67 & 88.89 & 20.63 & 23.70 & 31.25 & $3.43\times$ \\
        \rowcolor{weaverow}
        \textbf{WEAVE (Ours)} & \textbf{83.76} & \textbf{53.33}
        & \textbf{90.33} & \textbf{22.44} & \textbf{33.84} & \textbf{43.97}
        & $\mathbf{4.26\times}$ \\
        \midrule
        \multicolumn{8}{l}{\textbf{Kimi-Linear-48B}} \\[-2pt]
        Native         & 50.57 & 66.67 & 86.67 & 23.08 & 28.66 & 19.72 & $1.00\times$ \\
        INT8           & 48.81 & 63.33 & 84.80 & 22.68 & 25.40 & 18.05 & $1.85\times$ \\
        INT4           & 47.62 & 50.00 & 80.00 & 22.30 & 15.12 & 11.58 & $3.26\times$ \\
        \rowcolor{weaverow}
        \textbf{WEAVE (Ours)} & \textbf{49.21} & \textbf{66.67}
        & \textbf{85.30} & \textbf{22.83} & \textbf{27.98} & \textbf{19.12}
        & $\mathbf{4.12\times}$ \\
        \bottomrule
    \end{tabularx}
    \endgroup
\end{table}

\begin{figure}[t]
    \centering
    \includegraphics[width=\linewidth]{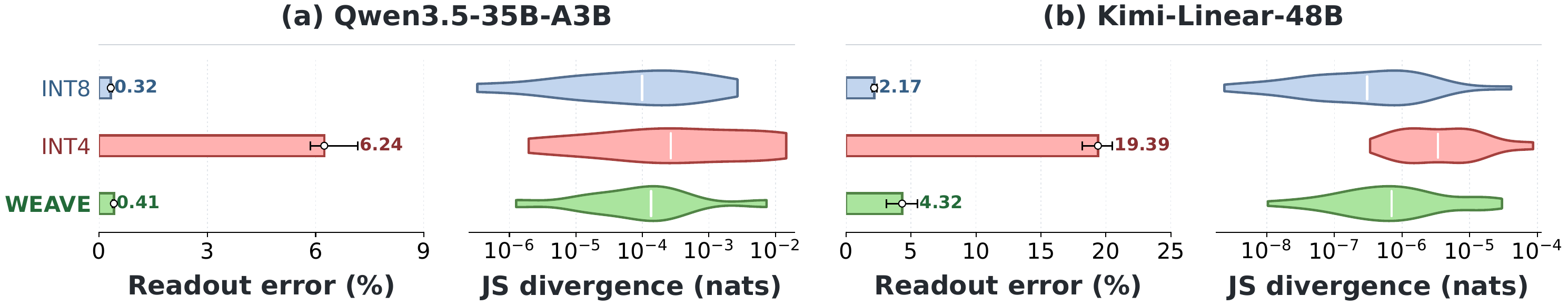}
    \caption{\textbf{Readout and output fidelity after restoration
    on AIME2026.}
    (a) Qwen; (b) Kimi.
    Left: median Native-query readout relative $\ell_2$ error with
    2.5th--97.5th percentile whiskers.
    Right: output JS distributions on a logarithmic axis.
    White markers denote medians.}
    \label{fig:restoration_error}
\end{figure}

\paragraph{Recurrent readout under native queries.}
Task scores show the eventual effect of compression but not how restoration
error first appears. We therefore replay native queries on restored recurrent
states and measure the relative error of the resulting readouts.
Figure~\ref{fig:restoration_error} shows that WEAVE stays close to INT8 and far below INT4 on both
models: on Qwen3.5-35B-A3B, WEAVE yields a readout error of 0.41\%,
compared with 0.32\% for INT8 and 6.24\% for INT4; on Kimi-Linear-48B,
the corresponding values are 4.32\%, 2.17\%, and 19.39\%.
These results indicate that WEAVE preserves recurrent-state usage much more
faithfully than aggressive low-bit quantization.

\paragraph{Full-vocabulary output divergence.}
To test whether state deviations affect predictions, we run the native and
restored states on the same teacher-forced suffix and compute the
Jensen--Shannon (JS) divergence between their full-vocabulary output
distributions at each step. Figure~\ref{fig:restoration_error} shows that WEAVE keeps output
divergence in the same low-divergence regime as INT8 and well below INT4 on
both models. The gap is especially clear on Kimi-Linear-48B, where WEAVE
remains close to INT8 while INT4 shifts to substantially larger divergence.
Together with the downstream task results, these measurements show that WEAVE
preserves both recurrent-state readout and next-token prediction behavior
after restoration.
%


\subsection{Online Serving Benefits}

We evaluate WEAVE in an end-to-end agent serving workload using SWE-bench
request streams with long, repeatedly reused prefixes. We replay identical
streams at increasing concurrency while fixing the scheduler, prefix-cache
policy, checkpoint placement, and GPU memory budget. The TTFT SLO is set to
$2\times$ the median isolated prefill time; SLO attainment is the fraction of requests
whose TTFT meets this threshold. Token goodput measures the aggregate
output-token throughput contributed by SLO-compliant requests. All runs include
checkpoint encoding and restoration. WEAVE keeps the same checkpoint density
but reduces their memory footprint, allowing more reusable prefixes to remain
in GPU memory. Under high load, this substantially reduces eviction and
recomputation: at concurrency 48, the dense baseline recomputes 68\% of
prefixes, while WEAVE converts the retained cache capacity into lower latency
and higher effective throughput.

\begin{figure}[t]
    \centering
    \includegraphics[width=\linewidth]{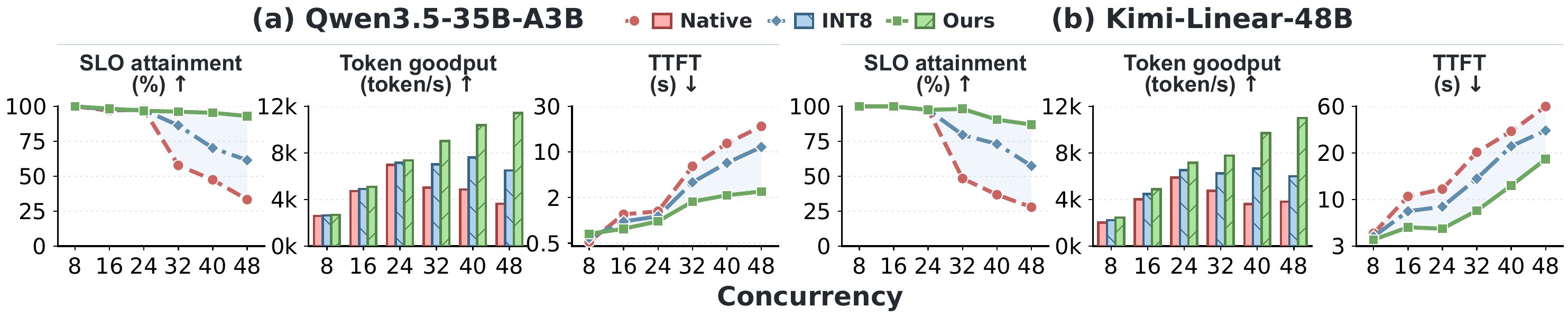}
    \caption{\textbf{End-to-end serving performance under increasing concurrency.}
    For (a) Qwen and (b) Kimi, we compare Native, INT8 and WEAVE (Ours)
    in terms of SLO attainment, token goodput, and time-to-first-token
    (TTFT) as concurrency increases.}
    \label{fig:serving}
\end{figure}

Figure~\ref{fig:serving} shows that these benefits grow rapidly with concurrency, with INT8 consistently falling between Native and WEAVE. At
concurrency 48, Qwen's TTFT SLO attainment improves from 33.3\% to 93.1\%,
while token goodput increases from 3,638 to 11,432 ($3.14\times$). Kimi shows
the same trend, with SLO attainment increasing from 27.9\% to 87.0\% and
token goodput from 3,824 to 10,995 ($2.88\times$). The reduction in
recomputation also directly improves latency: WEAVE reduces Qwen TTFT from
21.248\,s to 3.051\,s, an 85.6\% reduction ($6.96\times$ speedup), and Kimi
TTFT from 59.88\,s to 18.67\,s, a 68.8\% reduction ($3.21\times$ lower).
Overall, under increasing memory contention, WEAVE couples compact checkpoint
storage with efficient online compression and restoration, reducing TTFT and
substantially improving effective serving throughput together.


\subsection{Ablation Study}
\label{ablation}
\paragraph{Representation error and targeted repair.}
We sample checkpoints from MMLU at a 528-token offset to examine the
quality impact of replacing dense factorization with WEAVE's online
encoding pipeline (Table~\ref{tab:weave_ablations}(a)).
Dense residual SVD provides the low-rank reference with a
state-relative P95 error of 3.25\%.
Replacing it with recurrence-driven sketches avoids dense factorization
at the cost of a modest increase to 4.16\%.
Per-head rank allocation and factor packing then produce compact
checkpoints with a state-relative P95 error of 3.47\%.
Finally, targeted repair touches only seven of 960 heads and reduces
state-relative P95 to 3.38\%, while restoring the correct answer.
This shows that WEAVE can replace expensive dense factorization while
recovering the small accuracy loss through lightweight, targeted correction.

\begin{table}[t]
\centering
\caption{Representation error and online encoding cost.
Left: state-relative P95 reconstruction errors on MMLU
samples at a 528-token checkpoint offset.
WEAVE uses a fixed per-head rank allocation.
Right: online encoding time.}
\label{tab:weave_ablations}
\small
\setlength{\tabcolsep}{2pt}
\renewcommand{\arraystretch}{0.95}
\begin{minipage}[t]{0.49\linewidth}
\textbf{(a) Compression error}\par\vspace{2pt}
\begin{tabular*}{\linewidth}{@{\extracolsep{\fill}}lr@{}}
\toprule
Encoding configuration & P95 error \\
\midrule
Dense residual SVD
  & 3.25\% \\
+ Recurrence-driven sketch construction
  & 4.16\% \\
+ Per-head ranks and factor packing
  &  3.47\% \\
+ Targeted tail repair (WEAVE)
  &  3.38\% \\
\bottomrule
\end{tabular*}
\end{minipage}\hfill
\begin{minipage}[t]{0.49\linewidth}
\textbf{(b) Online encoding overhead}\par\vspace{2pt}
\begin{tabular*}{\linewidth}{@{\extracolsep{\fill}}lr@{}}
\toprule
Encoder path & Time (ms) \\
\midrule
Dense residual + SVD       & 995.16 \\
Native sketches + SVD      & 588.25 \\
WEAVE encoder (PyTorch)    &  69.01 \\
WEAVE encoder (Triton)     &  10.59 \\
\bottomrule
\end{tabular*}
\end{minipage}
\end{table}

\paragraph{Online encoding cost.}
In Table~\ref{tab:weave_ablations}(b), we measure checkpoint encoding overhead during the online model serving. Each measurement covers all 480 heads in one checkpoint; the
prior test in Section~2.2 evaluated only 32 heads. Constructing native sketches instead of a dense residual
reduces the SVD-based path from 995.16 to 588.25~ms
($1.69\times$), showing the benefit of avoiding the full matrix.
Replacing SVD with WEAVE's projected factor construction lowers
the cost to 69.01~ms, a further $8.52\times$ reduction. Grouped
Triton kernels cut the remaining small-operation and launch
overhead, reaching 10.59~ms and providing another $6.51\times$
speedup. Separately, load-gated asynchronous compaction introduces modest foreground
interference, with measured overheads of at most 3.35\% in batch wall time
and 3.51\% in tail latency.
Appendix~\ref{app:online_overhead} provides the detailed measurements and
experimental scope.


\section{Related Work}
\label{sec:related}

\subsection{KV Cache Compression for Full Attention}
A large body of work reduces the memory footprint of token-indexed KV
caches in full-attention models.
KIVI, KVQuant, and KVServe exploit numerical distributions, outliers, and
token importance to support low-bit KV representations
\citep{liu2024kivi,hooper2024kvquant,liu2026kvserve,wu2026beyond}.
More general quantization techniques, such as outlier isolation in SpQR
and rotation-based preprocessing in QuaRot, provide additional ways to
improve low-bit representations
\citep{dettmers2024spqr,ashkboos2024quarot}.
Other methods exploit structural redundancy.
Palu and Eigen Attention project KV representations into lower-dimensional
spaces, while ShadowKV uses the low-rank structure of keys to reduce GPU
memory consumption
\citep{chang2025palu,saxena2024eigen,sun2025shadowkv}.
A separate line of work retains only influential KV entries based on
attention patterns or observations collected before decoding
\citep{zhang2023h2o,li2024snapkv,cai2024pyramidkv,liu2026read}.
Recent work also explores lossless KV compression.
SplitZip exploits redundancy in floating-point exponents
\citep{guo2026splitzip}.
These KV compression methods target caches organized along the token
dimension.
WEAVE addresses fixed-size recurrent states and exploits correlations
between checkpoints along the same state trajectory.

\subsection{Recurrent-State Optimization for Linear Attention}
Recent work has begun to study the structure and serving cost of recurrent
states in linear-attention models.
State Rank Dynamics identifies stable rank differences across recurrent
heads and uses them to guide head-level pruning
\citep{sun2026rankdynamics}.
State Reduction instead removes key and query channels through structured
pruning, directly reducing the dimensions of each recurrent state
\citep{nazari2026statereduction}.
DASC exploits differences in state retention horizons to selectively
store recurrent heads or channels, recovering omitted components through
zero-filling or replay
\citep{yu2026dasc}.

Other work focuses on state reuse during serving.
Marconi improves prefix caching for hybrid LLMs through reuse-aware
admission and computation-aware eviction
\citep{pan2025marconi}.
Sparse Prefix Caching optimizes the placement of exact checkpoints and
recomputes tokens after the nearest cached state, while HYPIC composes states
from independently cached prompt segments
\citep{shirokikh2026sparseprefix,liu2026hypic}.

While existing compression methods reduce state size by selectively
discarding components or dimensions, WEAVE exploits redundancy across
checkpoints, using shared anchors and compact low-rank residuals to
preserve information across heads without storing their full states
independently.
WEAVE is also complementary to cache management policies, reducing the
storage cost of retained checkpoints rather than determining which
checkpoints to retain.


\section{Conclusion}
\label{sec:conclusion}
Hybrid linear attention shifts long-context serving from a growing KV cache
to fixed-size recurrent states, but agentic prefix reuse turns these states
into a persistent, random-access checkpoint store. WEAVE treats historical
checkpoints not as independent dense tensors, but as correlated points along
the model's native state trajectory. \emph{The key is to exploit this temporal structure without sacrificing
direct recovery of historical checkpoints.} It follows the recurrence to construct
shared-anchor residuals online, avoiding expensive post-hoc factorization,
while periodic compaction bounds recovery dependencies and system co-design
keeps construction and restoration practical for serving.
This allows randomly hit historical prefixes to recover compact states
directly, preserving model quality while increasing effective cache capacity
and reducing eviction and recomputation under load.

More broadly, WEAVE shows that model-state evolution can itself serve as a
compression primitive. By exploiting structure while state is being created,
rather than compressing it only after materialization, inference systems can
jointly optimize representation compactness, random recoverability, and
online execution cost. Looking forward, jointly managing recurrent-state
checkpoints and full-attention KV caches may enable a unified cache system
that allocates memory and compression effort across both representations
according to workload demand.

\bibliography{iclr2027_conference}
\bibliographystyle{iclr2027_conference}

\appendix
\section{Appendix}

\subsection{Encoding and Checkpoint Details}
\label{app:encoding_details}

Both model paths encode innovations as native updates arrive without forming
or factorizing a full $128\times128$ residual. GDN uses two-sided sketches;
KDA carries bounded, channel-aware factor pairs and solves only small
factor-space cores.

\paragraph{Qwen3.5 rank calibration and sketch dimensions.}
We calibrate the static per-head rank map from recurrent-state captures on four
long-context inputs: Code Repository Understanding (32K tokens),
Multi-Document QA (16K and 32K), and RepoQA (128K). For each of the 960 global
GDN heads, candidate ranks $\{8,12,16,24,32,48,64\}$ are evaluated on the
calibration residual grid. We take the largest relative Frobenius
reconstruction error over the grid for each input,
then its 95th percentile across inputs, and choose the smallest rank meeting
the $5\%$ target. The selected rank is rounded up to a stored-rank bucket
$r\in\{8,16,32,64\}$. This map is fixed before serving; TP=2 assigns 480
heads to each GPU.

For a $128\times128$ GDN state, let $w_r$ denote the working rank, and set
$q_r=w_r+8$ and $\ell_r=\min(2q_r+8,128)$. The fixed choices are
\begin{equation}
\begin{array}{c|rrrr}
r&8&16&32&64\\ \hline
w_r&16&32&48&80\\
q_r&24&40&56&88\\
\ell_r&56&88&120&128
\end{array}.
\label{eq:appendix_sketch_widths}
\end{equation}
For each head, $\Omega\in\mathbb{R}^{128\times q_r}$ and
$\Psi\in\mathbb{R}^{128\times\ell_r}$ are fixed seeded Gaussian matrices
orthonormalized by thin QR. The accumulator keeps
$Y=(H-H_a)\Omega\in\mathbb{R}^{128\times q_r}$ and
$Z=\Psi^\top(H-H_a)\in\mathbb{R}^{\ell_r\times128}$; these sketches and
projections are encoder workspace, not persistent checkpoint bytes. A native
chunk with scalar decay $d_j$ and write $A_jB_j^\top$ updates the fixed-anchor
residual by
\begin{equation}
R_{j+1}=d_jR_j+(d_j-1)H_a+A_jB_j^\top,
\qquad R_j:=H_j-H_a.
\label{eq:appendix_gdn_recurrence}
\end{equation}
Projecting this identity updates $Y$ and $Z$ without forming $R_j$. At a
compact boundary, two-pass shifted Cholesky QR of $Y$, a regularized projected
solve using $Z$, and two covariance subspace iterations produce the stored
rank-$r$ factors. Working ranks exceed stored ranks by 8--16 dimensions; only
the final factors are truncated and row-wise INT8 packed.

All calibration examples are disjoint from the evaluation
examples, and the rank map is fixed before evaluation
without further tuning.

\paragraph{Publication certificate, repair, and anchor span.}
The rank map is a capacity policy; publication additionally checks the
reconstructed state against the exact state already available at that
boundary. For head $h$ the certificate is
$e_h=\|H_h-\widehat H_h\|_F/\max(\|H_h\|_F,\epsilon)$.
Heads above the $5\%$ outer bound receive a dense BF16 correction
$E_h=H_h-\widehat H_h$; a head whose correction fails the $4\%$ safe
envelope is stored at exact native precision. Selective repair is bounded to
32 global heads per checkpoint. If the resulting checkpoint fails the
certificate, the exact checkpoint is retained. Thus the decision is made
while the exact state is available, not inferred during restore. Across the
tested checkpoints, the highest observed selective-repair share is seven of
960 heads ($0.729\%$), stored as BF16 correction tails; that checkpoint
requires neither an exact-head escape nor a whole-checkpoint exact fallback.
This figure denotes head-level repair, rather than the rate of whole-checkpoint
exact fallback.

We describe Qwen3.5 checkpoints with a 4224-token shared-anchor span and a
528-token child interval. Thus, each anchor has at most seven residual
children; the eighth interval starts the next anchor. The anchor holds native
BF16 recurrent state. Each child holds row-wise INT8 factor codes, FP16 row
scales, its BF16 convolution state, and any repair or exact-escape slabs.
The native transition chunk, prefix-hash unit, and cache page are separate
from this anchor-span geometry.

\begin{table*}[h]
\centering
\small
\setlength{\tabcolsep}{6pt}
\caption{Foreground interference of load-gated asynchronous compaction in the
isolated 8,192-token recomputation benchmark. Times are in seconds; increase
is relative to the run without background compaction.}
\label{tab:appendix_online_overhead}
\begin{tabular}{r rrr rrr}
\hline
& \multicolumn{3}{c}{Request TTFT P95 (s)}
& \multicolumn{3}{c}{Median batch wall time (s)} \\
Batch & No background & Async compaction & Increase
      & No background & Async compaction & Increase \\
\hline
1  & 0.34772 & 0.35506 & 2.11\% & 0.35581 & 0.35727 & 0.41\% \\
8  & 2.37314 & 2.41170 & 1.62\% & 2.39024 & 2.42866 & 1.61\% \\
16 & 4.68269 & 4.84693 & 3.51\% & 4.72709 & 4.88551 & 3.35\% \\
32 & 9.29379 & 9.54691 & 2.72\% & 9.42870 & 9.68205 & 2.69\% \\
\hline
\end{tabular}
\end{table*}

\paragraph{Period-two compaction.}
For the 4224-token anchor span with 528-token child spacing, period two counts
child publication points rather than native chunks. Compact checkpoints are
materialized every two child boundaries; the intermediate boundary keeps a
bounded one-step dependency on the latest compact checkpoint. Thus, recovery
depth is at most one while compaction is amortized across two child
publications. Native 64-token chunks are only the internal update granularity.
An intermediate checkpoint stores only a bounded transition from the latest
compact checkpoint: the accumulated decay $d$, a fixed-rank factor pair
$L,K$, their packing metadata, and the current convolution state.
Recovery applies
\begin{equation}
\widehat H_{\mathrm{int}}
=
d\,\widehat H_{\mathrm{compact}} + LK^\top .
\end{equation}
The referenced compact checkpoint is retained while any dependent intermediate
checkpoint remains live, so the recovery depth is at most one.
All intermediate payloads are included in the physical checkpoint accounting
reported in Table~\ref{tab:quality}.

\paragraph{KDA state convention and bounded encoder.}
\label{app:kda}
KDA uses a channel-transport specialization. For each TP-local head,
$H_t\in\mathbb{R}^{d_v\times d_k}$ with $d_v=d_k=128$,
$v_t\in\mathbb{R}^{d_v}$, $k_t\in\mathbb{R}^{d_k}$,
$D_t=\operatorname{diag}(d_t)\in\mathbb{R}^{d_k\times d_k}$, and
$W_t\in\mathbb{R}^{d_v\times d_k}$. The native layout stores the key/channel
dimension last, so decay multiplies on the \emph{right}:
$H_{t+1}=H_tD_t+W_t$. A native chunk supplies effective write factors
$L_j,K_j\in\mathbb{R}^{128\times s_j}$, $s_j\leq64$, with its internal gate
weighting already applied. Define the accumulated diagonal transport
$P_{a:e}=\prod_{j=a}^{e-1}D_j$. The endpoint innovation is represented as
$R_{a:e}\approx F_eG_e^\top$ without constructing $R_{a:e}$ densely.

\paragraph{Error propagation.}
Let $E_t = R_t - \widehat{R}_t$ denote the difference between
the exact and approximate residuals driven by the same native
updates. For KDA, the encoding error evolves as
\begin{equation}
    E_{t+1} = E_t D_t + \delta_{t+1},
\end{equation}
where $\delta_{t+1}$ denotes the local truncation error.
Assuming $\|D_t\|_2 \leq \rho < 1$ and
$\|\delta_t\|_F \leq \varepsilon$, repeated application of
the norm inequality gives
\begin{equation}
    \|E_t\|_F \leq
    \rho^t \|E_0\|_F +
    \frac{\varepsilon(1-\rho^t)}{1-\rho}.
\end{equation}
Thus, under uniform contraction, the accumulated encoding
error is uniformly bounded over time. In the non-expansive
case ($\rho=1$), the bound becomes
\begin{equation}
    \|E_t\|_F \leq \|E_0\|_F + t\varepsilon.
\end{equation}
Quantization introduces additional reconstruction error,
which is evaluated separately through checkpoint validation
and repair.

\noindent\fbox{\begin{minipage}{0.94\linewidth}
\textbf{KDA bounded factor encoding (per head).} For stored rank $r\in\{8,16,32,64\}$, selected by the same fixed per-head
rank policy as GDN, use working width $w=r+8$. Initialize $p\leftarrow\mathbf{1}\in\mathbb{R}^{128}$
and empty factors $F,G\in\mathbb{R}^{128\times0}$. For each native chunk $j$:
\begin{enumerate}
\setlength{\itemsep}{0pt}
\item Read channel decay $d_j$ and effective write factors $L_j,K_j$.
\item Set $p\leftarrow p\odot d_j$,
      $\widetilde F=[F,L_j]$, and
      $\widetilde G=[\operatorname{diag}(d_j)G,K_j]$.
\item Apply thin QR to $\widetilde F,\widetilde G$; truncate the SVD of
      their factor-space core to width $w$, yielding updated $F,G$.
\end{enumerate}
At endpoint $e$, truncate the factor-space core to the head's stored rank
$r$, form balanced factors $A_e,B_e$, row-wise INT8-pack both, and store
BF16 $p$.
\end{minipage}}

The identity
$FG^\top\operatorname{diag}(d_j)+L_jK_j^\top
=[F,L_j][\operatorname{diag}(d_j)G,K_j]^\top$
shows that channel-wise transport acts on the right factor. Each update
temporarily has at most $w+64$ columns, and all truncation is confined to
the corresponding factor-space core; no full residual matrix or full-state
SVD is needed. Approximation enters when this core is truncated, and the
publication certificate checks the final packed candidate. The checkpoint
stores an anchor ID,
boundary and TP identity, the BF16 accumulated transport vector $p$, the
INT8 factors and their FP16 row scales, and the endpoint BF16 convolution
state. With the separately stored row-wise INT8 anchor, restoration is
\begin{equation}
\widehat H_e=Q_8(H_a)\operatorname{diag}(p_{a:e})
+Q_8(A_e)Q_8(B_e)^\top.
\label{eq:appendix_kda_restore}
\end{equation}
The transport vector is persisted with the endpoint; no intermediate
checkpoints are replayed. The KDA quality evaluation uses a 6144-token
anchor-to-endpoint span.

\subsection{Online Overhead Test}
\label{app:online_overhead}

We measured the foreground interference of load-gated asynchronous checkpoint
compaction using an isolated 8,192-token recomputation benchmark on
Qwen3.5-35B-A3B with TP2. Across batch sizes of 1, 8, 16, and 32, adding
background compaction increased median batch wall time by 0.41\%--3.35\% and
request TTFT P95 by 1.62\%--3.51\% relative to runs without background
compaction. This benchmark intentionally removes cache-capacity benefits and isolates only
the foreground interference introduced by asynchronous background compaction
under the same recomputation workload. The largest observed increase is
3.35\% in median batch wall time and 3.51\% in TTFT P95.
These numbers therefore characterize contention from the asynchronous
compaction path on Qwen, rather than the end-to-end overhead of the complete
codec.

\paragraph{Full codec overhead.}
Table~\ref{tab:full_codec_overhead} reports the complete
time and memory costs of WEAVE on both models, including
encoding, restoration, certification, repair, and fallback.
Codec timings are median GPU-active times measured on
MMLU-Pro using the checkpoint configuration described
in Appendix~\ref{app:encoding_details}.

Online overheads are measured against exact checkpoints
at concurrency 16, including recomputation caused by
fallback events. The full codec introduces only
0.7--1.6\% JCT overhead and 2.6--3.4\% TTFT overhead,
which are small relative to the serving gains under
memory pressure.

\begin{table}[htbp]
\centering
\small
\caption{Full codec time and memory overhead.
Persistent storage is aggregated across two GPUs,
while peak workspace is the maximum per GPU.}
\label{tab:full_codec_overhead}
\begin{tabular}{lrr}
\toprule
\textbf{Metric} & \textbf{Qwen3.5-35B-A3B}
& \textbf{Kimi-Linear-48B} \\
\midrule
Persistent storage (GiB) & 0.63 & 0.398 \\
Peak workspace (MiB)    & 358 & 212 \\
Restore time (ms)       & 0.234 & 0.189 \\
Full codec time (ms)    & 18.50 & 9.96 \\
JCT overhead            & 1.60\% & 0.70\% \\
TTFT overhead           & 3.40\% & 2.60\% \\
\bottomrule
\end{tabular}
\end{table}

\subsection{Quality-evaluation protocol.}
All downstream quality results use the full official evaluation set of each
benchmark rather than a sampled subset. Each task is evaluated in three
independent runs, and Table~\ref{tab:quality} reports the mean score across
runs. For repeated-restoration evaluation, checkpoints follow the same fixed
placement schedule used throughout the experiment. Each trajectory therefore
contains 39--157 restore-and-continue events (96 on average), depending on its
length; every restored state is subsequently used to continue inference and
construct later checkpoints.
 \subsection{Serving Workload and Settings}
\label{app:serving_setup}

\paragraph{Workload construction.}
Inspired by the prefix-reuse patterns in AgentX
\citep{agentx_trace}, we construct a workload of
240 replay requests from 48 agent sessions,
including 48 warm-up requests and 192 measured
requests. Sessions start at two-second intervals,
while request intervals within each session follow
the original trace.

\paragraph{Cache configuration.}
The combined GPU memory budget for recurrent states
and full-attention KV caches is 46.27~GiB per GPU.
All configurations use the same request stream,
arrival pattern, and memory budget.

\paragraph{Measurement protocol.}
Each configuration is replayed three times, and
the reported results are averaged across runs.
TTFT is measured on the server side, and all
serving measurements include checkpoint encoding
and restoration. Warm-up requests are excluded
from the reported results.

\end{document}